\documentclass[10pt,twocolumn,letterpaper]{article}
\usepackage[pagenumbers]{cvpr}      %
\usepackage{cuted} %
\usepackage{currfile} %
\usepackage{caption} %

\usepackage{multirow}

\usepackage{listings}
\lstdefinestyle{prompt}{
  basicstyle=\ttfamily\scriptsize,
  breaklines=true,
  breakatwhitespace=false,
  columns=fullflexible,
  keepspaces=true,
  showstringspaces=false,
  frame=single,
  rulecolor=\color{black!45},
  backgroundcolor=\color{black!4},
  aboveskip=6pt,
  belowskip=10pt,
  xleftmargin=2pt,
  xrightmargin=2pt,
  framexleftmargin=4pt,
  framexrightmargin=4pt,
  framextopmargin=4pt,
  framexbottommargin=4pt,
}
\newcommand{\promptlisting}[2]{%
  \par\addvspace{0.8em}%
  \noindent
  \colorbox{black!12}{%
    \parbox{\dimexpr\linewidth-2\fboxsep\relax}{\raggedright\bfseries\footnotesize #1}%
  }%
  \par\nopagebreak\vspace{6pt}%
  \lstinputlisting[style=prompt]{#2}%
}
\def\figref#1{figure~\ref{#1}}

\def\secref#1{section~\ref{#1}}

\definecolor{cvprblue}{rgb}{0.21,0.49,0.74}
\usepackage[pagebackref,breaklinks,colorlinks,allcolors=cvprblue]{hyperref}
\def\confName{CVPR}
\def\confYear{2027}
\title{Adaptive-Group Process Rewards for GUI Agents}
\author{%
\normalsize Zirui Shang$^{1,2}$, Yang Liu$^{2}$, Xin Shu$^{2,3}$, Zhi Gao$^{1,2,4\dagger}$, Xinxiao Wu$^{1,4\dagger}$, Lifeng Fan$^{2}$\\[4pt]
\small $^1$Beijing Key Laboratory of Intelligent Information Technology,\\
\small School of Computer Science \& Technology, Beijing Institute of Technology\\
\small $^2$State Key Laboratory of General Artificial Intelligence, BIGAI\\
\small $^3$Wuhan University\\
\small $^4$Guangdong Laboratory of Machine Perception and Intelligent Computing,\\
\small Shenzhen MSU-BIT University
}
\hypersetup{pdftitle={Adaptive-Group Process Rewards for GUI Agents},
  pdfauthor={Zirui Shang, Yang Liu, Xin Shu, Zhi Gao, Xinxiao Wu, Lifeng Fan}}
\begin{document}
\maketitle
\begingroup
\renewcommand{\thefootnote}{\fnsymbol{footnote}}
\footnotetext[2]{Corresponding authors.}
\endgroup
\begin{abstract}
Training graphical user interface (GUI) agents with reinforcement learning suffers from sparse outcome rewards, leading to inefficient learning.
Process rewards offer a promising direction for credit assignment to intermediate actions; however, accurately evaluating an action requires identifying evidence across long-horizon GUI trajectories, as its value may depend on information beyond its immediate state and observation.
In this paper, we propose an adaptive-group process reward framework that dynamically partitions a GUI trajectory into semantically coherent subgoal groups, enabling process reward estimation at both the action and group levels.
At the action level, a critic evaluates each action based on the visual changes and subsequent actions within the group for action correctness and contribution to the subgoal.
At the group level, the framework aggregates action-level evaluations to assess subgoal completion and contribution to the overall task, and further refines these assessments by leveraging the context of all groups in the trajectory.
The process reward of each action is obtained by combining its action-level evaluation with the refined group-level evaluation.
We further augment GRPO's outcome advantage with the proposed process rewards to provide finer-grained guidance for GUI agent training.
Experiments on process reward benchmarks and downstream RL training demonstrate the effectiveness of the proposed framework.
\end{abstract}
\section{Introduction}
\label{sec:intro}
\begin{figure*}[t]
\centering
\includegraphics[width=0.9\textwidth]{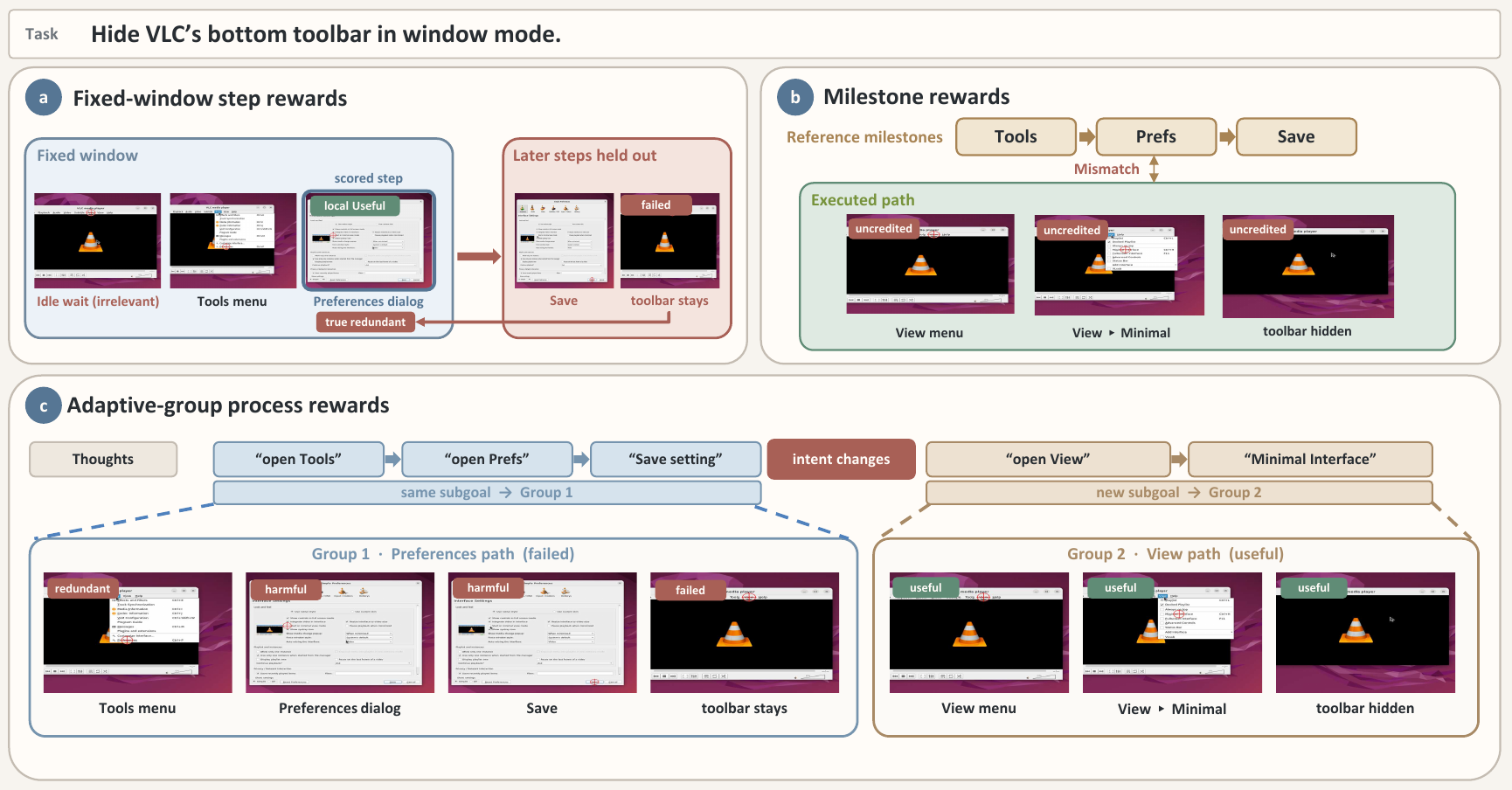}
\caption{Fixed-window step rewards, milestone rewards, and adaptive-group process rewards on an executed GUI trajectory. (a)~A fixed local window can include irrelevant context and miss later evidence. (b)~Reference milestones can leave a valid executed path uncredited. (c)~Adaptive groups follow the executed subgoals and score actions within each group.}
\label{fig:intro}
\end{figure*}

Recent advances in large language models and vision-language models have enabled graphical user interface (GUI) agents~\cite{cheng2024seeclick,hong2024cogagent,qin2025ui,lin2024showui,xu2024aguvis,gao2024multi,zhang2025tongui}. These agents translate user instructions and visual observations into executable interactions across digital devices, with applications such as software testing, office assistance, and industrial automation. GUI tasks typically involve long-horizon, multi-turn interactions~\cite{zhou2023webarena,rawles2024androidworld,xie2025osworld}, requiring agents to maintain multimodal interaction context across changing visual states and coordinate sequences of actions. Reinforcement learning (RL)~\cite{li2025dart,xiao2025uigenie,lu2025orcust,tang2026clawgui} is well suited to such interactive tasks, allowing agents to improve their policies based on feedback from the GUI environments.
Most GUI environments provide reward signals primarily in the form of sparse outcome rewards~\cite{xie2025osworld,li2025dart}: a trajectory is judged successful if the final application state satisfies the instruction requirement, and unsuccessful otherwise. Consequently, the outcome-level signal is assigned to all intermediate actions in a trajectory, failing to distinguish useful actions from redundant or incorrect ones and making credit assignment unreliable. Process rewards~\cite{lightman2023verify,wang2023mathshepherd,chen2025guishepherd} are therefore needed to evaluate intermediate actions. However, constructing reliable process rewards for GUI trajectories is challenging because GUI environments rarely provide direct supervision for intermediate actions. Whether an action is correctly executed can often be verified from the current and subsequent observations, whereas whether its semantic purpose is correct may only become clear after subsequent actions. Judging the action requires evidence from a broader trajectory context than the immediate state change after that action alone.
Representative GUI process reward methods organize trajectory evidence and assign credit through two approaches: fixed-window step rewards~\cite{chen2025guishepherd,xiao2025uigenie,lu2025orcust,wan2026seekjudge} and milestone rewards~\cite{zheng2026admire,zhang2025progrm}. Fixed-window step rewards evaluate an action within a fixed local context (\figref{fig:intro}(a)). Short windows may miss long-range evidence needed to assess the action's effect, whereas longer windows may introduce irrelevant context. Milestone rewards instead define task progress around reference checkpoints or subgoals (\figref{fig:intro}(b)). Because GUI tasks may admit multiple valid execution paths, such milestones can under-credit progress that does not follow the reference milestone sequence and overlook the intermediate actions connecting checkpoints. Thus, an effective process reward should adapt its evidence scope and credit assignment to the executed trajectory, rather than relying on fixed local scopes or reference milestone sequences.
In this paper, we introduce an adaptive-group process reward framework for GUI trajectories (\figref{fig:intro}(c)). Using the task instruction and the agent's responses, the executed trajectory is dynamically partitioned into consecutive groups of actions that share the same subgoal or semantic operation, rather than by a fixed partition rule. This grouping defines the context for judging the actions in each group and supports two complementary levels of evaluation. At the action level, the critic assesses whether each action is correctly executed based on the local visual change and whether it advances the group's subgoal based on later actions in the group.
At the group level, the critic aggregates the action-level assessments within each group to assess its subgoal completion and contribution to the whole task. Then, the produced group-level assessments are revisited in the context of all groups, allowing the critic to judge each group's role beyond its local evidence. The process reward of each action is obtained by combining its action-level assessment with the refined group-level assessment of its group.
We further use the generated process rewards to shape the GRPO~\cite{shao2024deepseekmath} advantage at each step, providing finer-grained credit assignment during policy optimization.
We evaluate the framework on process-reward benchmarks and in downstream RL on OSWorld. The results show consistent improvements in process-reward accuracy and task success rate, demonstrating the effectiveness of adaptive-group process rewards.
Our contributions are as follows:
\begin{itemize}
    \item We introduce an adaptive-group process reward framework for GUI agents, which dynamically partitions trajectories into semantically coherent subgoal groups to organize the evidence needed for process reward evaluation.
    \item We design a two-level evaluation that assesses execution and contribution at both the action level and the group level, and obtain the process reward of each action by combining these assessments, making correct execution and group-level contribution jointly necessary.
    \item Experiments on process reward benchmarks and downstream RL training demonstrate the effectiveness of the proposed framework.
\end{itemize}
\section{Related Work}
\label{sec:related}
\subsection{GUI Agents}
Recent progress in large language models and vision-language models has enabled GUI agents that interact with applications through visual observations and executable actions. Early systems mainly relied on prompting or supervised fine-tuning, while recent open-source models and frameworks, such as ShowUI~\cite{lin2024showui}, CogAgent~\cite{hong2024cogagent}, AGUVIS~\cite{xu2024aguvis}, UI-TARS~\cite{qin2025ui}, OpenCUA~\cite{wang2025opencuaopenfoundationscomputeruse}, Agent S~\cite{agashe2024agent}, and TongUI~\cite{zhang2025tongui}, improve grounding, planning, and cross-platform interaction through larger GUI data and stronger vision-language-action modeling. More recently, reinforcement learning has become a central direction for GUI agents because it can optimize long-horizon interaction from environment feedback rather than only imitating fixed demonstrations. DART~\cite{li2025dart} decouples multi-turn rollout from training for GUI RL, UI-Genie~\cite{xiao2025uigenie} builds a reward-guided self-improvement loop for mobile GUI agents, Orcust~\cite{lu2025orcust} combines principle-constrained reward modeling with online VM-grounded trajectory construction, and ClawGUI~\cite{tang2026clawgui} provides a full-stack infrastructure for GUI agent RL and deployment. However, most RL pipelines still rely heavily on sparse outcome rewards, which makes it difficult to assign credit to intermediate actions. Our work addresses this limitation by constructing dense process rewards over adaptive groups and the actions in those groups.
\subsection{Process Reward Modeling}
Process reward modeling supplements sparse outcome rewards with feedback on intermediate steps, enabling more fine-grained credit assignment during training. In reasoning~\cite{lightman2023verify,wang2023mathshepherd} and coding~\cite{le2022coderl,liu2023rltf,dai2026gvpo} tasks, intermediate feedback can often be verified from execution results, compiler errors, unit tests, or other deterministic checks. GUI tasks provide no comparable automatic signal for most intermediate operations: the value of a click, text entry, menu selection, or file operation depends on visual state changes, task requirements, and subsequent operations. GUI process reward modeling must therefore infer both whether an action is correctly executed and its semantic role from trajectory context.
Representative GUI process reward methods include fixed-window step rewards~\cite{chen2025guishepherd,xiao2025uigenie,lu2025orcust,wan2026seekjudge} and milestone rewards~\cite{zheng2026admire,zhang2025progrm}. Fixed-window step rewards judge actions within a fixed local context. GUI-Shepherd~\cite{chen2025guishepherd} learns step-level scores and rationales from human annotations, while SeekJudge~\cite{wan2026seekjudge} further targets fine-grained step-level judgments for use within reinforcement learning. GUI-PRA~\cite{xiong2025guipra} selects historical context and actively perceives UI state changes to improve step-level judgment. Milestone rewards organize feedback around reference checkpoints or subgoals. ADMIRE~\cite{zheng2026admire} distills verifiable milestones from successful explorations and applies asymmetric credit assignment to successful and failed trajectories. ProgRM~\cite{zhang2025progrm} extracts reference key steps from successful trajectories to construct progress labels for reward model training.
Beyond these approaches, StainFlow~\cite{hao2026stainflow} organizes progress evidence around tracked entity states and verified key nodes, whereas our framework organizes evaluation around the subgoals followed in the executed trajectory. GUIDE~\cite{zhai2026guide} segments completed trajectories for task-level evaluation. Our framework uses executed subgoal groups as contexts for per-action credit assignment. It assesses execution and contribution at both action and group levels, refines group-level judgments using the full group sequence, and combines action-level scores with refined group-level scores to produce process rewards for reinforcement learning.
\section{Method}
\label{sec:method}
\subsection{Problem Formulation}
\label{sec:problem-formulation}
\begin{figure*}[t]
\centering
\includegraphics[width=0.95\linewidth]{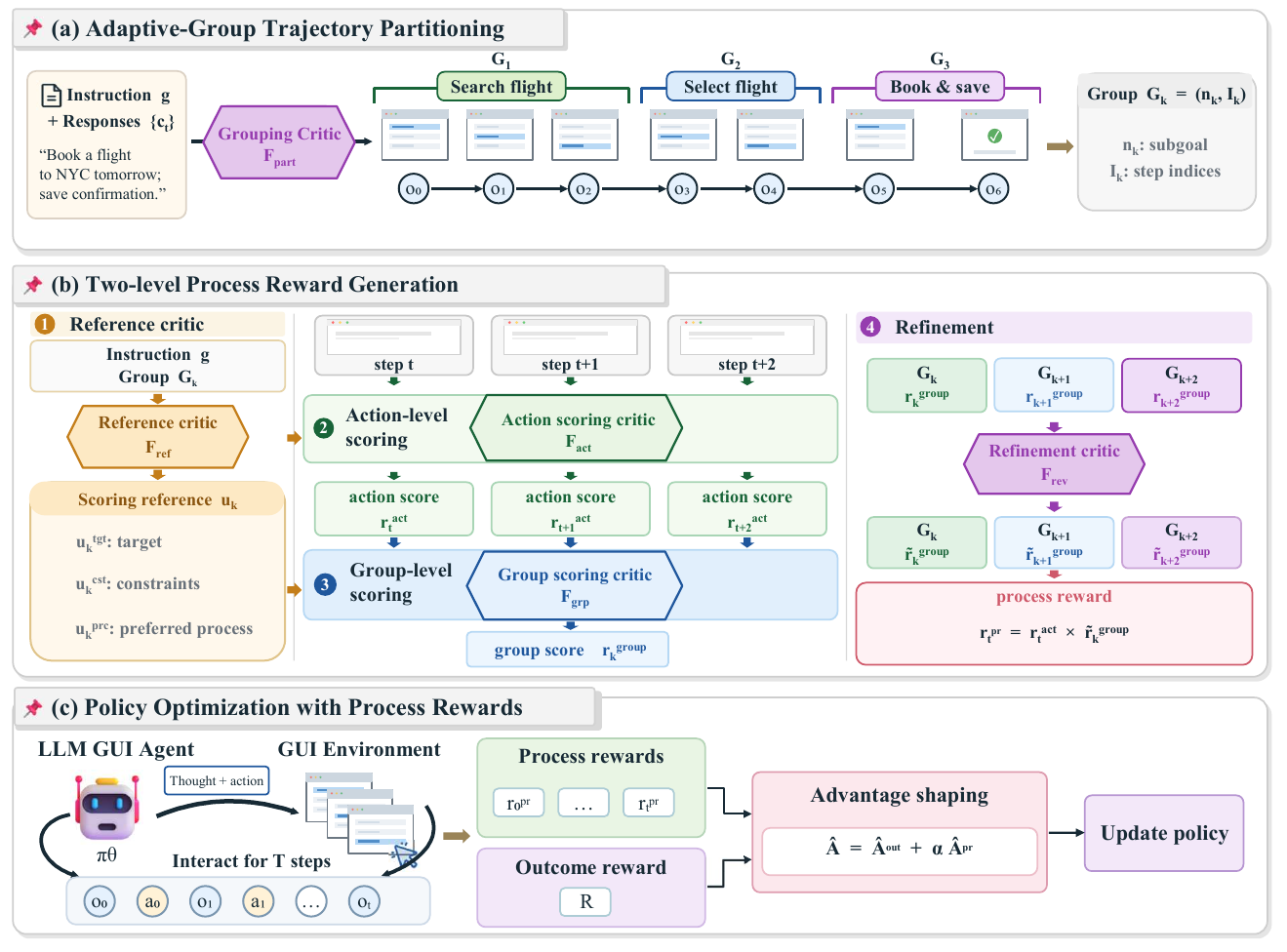}
\caption{Overview of the proposed framework. (a)~A grouping critic partitions the trajectory from the instruction and responses into subgoal groups. (b)~Two-level scoring produces an action-level score and a refined group-level score, which are used to generate the process rewards. (c)~The process rewards shape the GRPO advantage.}
\label{fig:framework}
\end{figure*}

We formulate solving a GUI task as a sequential interaction between an agent and an environment. Given a task instruction $g$, the agent observes the GUI state $o_t$ at each step. Conditioned on this observation and the interaction history, it generates a response $c_t$ with an executable action $a_t$. Executing $a_t$ changes the environment and produces the next observation $o_{t+1}$. An executed trajectory is therefore represented as
\begin{equation}
    \tau=(o_0,c_0,a_0,o_1,c_1,a_1,\ldots,o_{T-1},c_{T-1},a_{T-1},o_T),
\end{equation}
where $T$ is the number of executed actions. The response $c_t$ provides the agent's textual intent, while the observations and executed actions record the interaction with GUI.
After the trajectory is completed, the environment or a task verifier $V$ produces an outcome reward $R=V(g,\tau)$. This trajectory-level signal does not directly indicate the value of individual actions. We therefore formulate GUI process reward generation as producing a process reward for every executed action:
\begin{equation}
    (r_0^{\mathrm{pr}},r_1^{\mathrm{pr}},\ldots,r_{T-1}^{\mathrm{pr}})
    =\phi(g,\tau),
\end{equation}
where $\phi$ is a process reward model and $r_t^{\mathrm{pr}}$ is the process reward of action $a_t$ at step $t$. As shown in \figref{fig:framework}, we instantiate $\phi$ by first partitioning $\tau$ into adaptive groups that define the context for judging each action (\secref{sec:trajectory-partitioning}), then generating action-level and group-level assessments (\secref{sec:process-reward-generation}). The resulting process rewards are used to shape GRPO's step-level advantage during policy optimization (\secref{sec:pr-grpo}).
\subsection{Adaptive-Group Trajectory Partitioning}
\label{sec:trajectory-partitioning}
Fixed-window step rewards evaluate an action within a fixed local context, while milestone rewards organize credit around reference checkpoints. To define judgment contexts for per-action credit assignment, we recover subgoals from the executed trajectory instead. Grouping should follow executed intents, which are already stated in each $c_t$, so we use a grouping critic $F_{\mathrm{part}}$ to partition the trajectory from the instruction $g$ and the responses $\{c_t\}_{t=0}^{T-1}$:
\begin{equation}
    \{G_k\}_{k=1}^{K}
    = F_{\mathrm{part}}(g, \{c_t\}_{t=0}^{T-1}),
    \qquad
    G_k=(n_k,\mathcal{I}_k).
\end{equation}
The result is a contiguous, non-overlapping partition of the trajectory: each $G_k$ consists of a hypothesized subgoal $n_k$ and the step indices $\mathcal{I}_k$ of the group, and $k(t)$ denotes the unique group of step $t$. The subgoal $n_k$ describes the group's attempted intent rather than a required milestone of $g$, so different executions can induce different groups. These groups are the context for the judgments below.
\subsection{Adaptive-Group Process Reward Generation}
\label{sec:process-reward-generation}
The adaptive group defines the context for judging its actions and supports two complementary levels of evaluation: how an action contributes to the group's subgoal, and how the subgoal contributes to the overall task. At the action level, the critic assesses whether each action is correctly executed and whether it advances the group's subgoal. At the group level, it aggregates these assessments to judge the group's subgoal completion and contribution to the global task, then revisits the group-level assessments using all groups as context. Verifying execution can be done from the current and next observations, whereas judging semantic purpose depends on what follows: later actions in the group for an action, and other groups for a subgoal.
To score an action or a group, the executed actions have to be interpreted against the instruction. We therefore first use a reference critic $F_{\mathrm{ref}}$ to build, for each $G_k$, an instruction-conditioned reference $u_k$ from $g$, $G_k$, and the group's responses $\{c_t\}_{t\in\mathcal{I}_k}$, to inform the two levels of scoring that follow:
\begin{equation}
    u_k
    = (u_k^{\mathrm{tgt}}, u_k^{\mathrm{cst}}, u_k^{\mathrm{prc}})
    = F_{\mathrm{ref}}(g, G_k, \{c_t\}_{t\in\mathcal{I}_k}).
\end{equation}
$u_k^{\mathrm{tgt}}$ is the target state implied by $g$ for this group, $u_k^{\mathrm{cst}}$ the checkable constraints, and $u_k^{\mathrm{prc}}$ the preferred process. Action-level scoring consults $u_k^{\mathrm{cst}}$ when checking whether the change from $o_t$ to $o_{t+1}$ is consistent with the instruction; group-level scoring consults $u_k^{\mathrm{tgt}}$ and $u_k^{\mathrm{prc}}$ when judging whether the group advances the task and whether the attempted route is reasonable. We treat $u_k$ as a scoring reference rather than as a strict prior: the trajectory is not required to follow $u_k^{\mathrm{prc}}$ or to attain $u_k^{\mathrm{tgt}}$, and an alternative path that still satisfies $g$ remains valid.
\paragraph{Action-level scoring.}
The actions of $G_k$ are scored together, not one call per step. An action-level scoring critic $F_{\mathrm{act}}$ takes $g$, $u_k$, $G_k$, the group's responses $\{c_t\}_{t\in\mathcal{I}_k}$, and the group's screenshots, and returns a judgment for every action in $G_k$:
\begin{equation}
\begin{aligned}
    J_k & = \{(e_t, p_t, h_t)\}_{t \in \mathcal{I}_k} \\
        & = F_{\mathrm{act}}\bigl(g, u_k, G_k, \{c_t\}_{t \in \mathcal{I}_k}, \{o_t\}_{t \in \mathcal{I}_k\cup\{t_k^+\}}\bigr),
\end{aligned}
\end{equation}
where $t_k^+=1+\max\mathcal{I}_k$ is the observation after the last action of the group. The execution result $e_t$ takes one of four labels: Completed, Partial, Failed, or Uncertain. The contribution $p_t$ to the group's subgoal takes one of six labels: Milestone, Useful-efficient, Useful-inefficient, Exploratory, Redundant, or Harmful. Here, $h_t$ is a concise evidence summary of the action.
Later actions in $G_k$ remain available as post-action context, so $p_t$ can be judged from the subsequent semantic effect rather than from the local state change alone.
When the group is long and image-heavy, attaching every screenshot can exceed the context budget. We then split $\mathcal{I}_k$ into successive windows $\mathcal{W}\subseteq\mathcal{I}_k$ and apply the same critic to each window: the call still receives the whole group's responses as text, but screenshots only for $\{o_t\}_{t\in\mathcal{W}\cup\{t_{\mathcal{W}}^+\}}$, where $t_{\mathcal{W}}^+=1+\max\mathcal{W}$. Concatenating the window outputs yields $J_k$.
The two labels of each action are then mapped by a fixed table $M$ to a scalar action-level score $r_t^{\mathrm{act}}=M(e_t,p_t)$ in $\{0,0.2,0.4,0.6,0.8,1.0\}$ (Appendix~\ref{app:score-map}). The structured judgments $J_k$ are passed to group-level scoring.
\paragraph{Group-level scoring.}
Whether $G_k$ completes its subgoal, and whether the subgoal helps $g$, cannot be read from any single action. A group-level scoring critic $F_{\mathrm{grp}}$ therefore scores the group from the structured action-level judgments $J_k$---not the scalars $r_t^{\mathrm{act}}$ alone---together with $u_k$ and summaries $H_{<k}$ of preceding groups, which situate $G_k$ in the progress already made. It judges the group by $e_k$ for subgoal completion and $p_k$ for contribution to the task, then maps these labels with the same table $M$ to a group-level score $r_k^{\mathrm{group}}$:
\begin{equation}
\begin{aligned}
    (e_k, p_k, h_k) & = F_{\mathrm{grp}}(g, u_k, G_k, J_k, H_{<k}), \\
    r_k^{\mathrm{group}} & = M(e_k, p_k).
\end{aligned}
\end{equation}
$h_k$ is a concise summary of $G_k$ and is appended to $H_{<k}$ for later groups. Groups are scored in execution order, once per $G_k$.
\paragraph{Group-level refinement.}
After the initial pass scores groups in execution order, we use a refinement critic $F_{\mathrm{rev}}$ to revisit $(e_k,p_k)$ with $g$ and the full sequence of group conclusions, using this more complete context to reassess each group's role, correct potential group-level misjudgments, and obtain a refined group-level score $\tilde{r}_k^{\mathrm{group}}$:
\begin{equation}
\begin{aligned}
    \{(\tilde{e}_k, \tilde{p}_k)\}_{k=1}^{K}
    & = F_{\mathrm{rev}}\bigl(g, \{G_k, J_k, e_k, p_k\}_{k=1}^{K}\bigr), \\
    \tilde{r}_k^{\mathrm{group}} & = M(\tilde{e}_k, \tilde{p}_k).
\end{aligned}
\end{equation}
\paragraph{Process reward aggregation.}
The process reward of action $a_t$ uses the refined group-level score of $G_{k(t)}$ to modulate its action-level score $r_t^{\mathrm{act}}$:
\begin{equation}
\begin{aligned}
    r_t^{\mathrm{pr}}
    &= r_t^{\mathrm{act}}\left[\epsilon_{\mathrm{g}}+(1-\epsilon_{\mathrm{g}})\tilde{r}_{k(t)}^{\mathrm{group}}\right].
\end{aligned}
\end{equation}
High process rewards require both a high action-level score and a high group-level score. The factor $\epsilon_{\mathrm{g}}$ sets a lower bound on group-level scaling, preventing group-level judgments from entirely suppressing action-level credit, while preserving the constraint $r_t^{\mathrm{act}}=0\Rightarrow r_t^{\mathrm{pr}}=0$.
The score mapping $M$ and the critic prompts are given in Appendix~\ref{app:impl}.
\subsection{Policy Optimization with Process Rewards}
\label{sec:pr-grpo}
The process rewards $r_t^{\mathrm{pr}}$ are used to shape GRPO's advantage at each step. For an instruction $g$, GRPO~\cite{shao2024deepseekmath} samples $N$ trajectories $\mathcal{B}_g=\{\tau_i\}_{i=1}^{N}$ and standardizes their outcome rewards $\{R_i\}$ within $\mathcal{B}_g$ by the mean $\mu^{\mathrm{out}}$ and standard deviation $\sigma^{\mathrm{out}}$:
\begin{equation}
    \hat{A}_i^{\mathrm{out}}
    = \frac{R_i-\mu^{\mathrm{out}}}{\sigma^{\mathrm{out}}+\epsilon}.
\end{equation}
Here, $\epsilon>0$ is a small constant for numerical stability. This term is constant across the steps of $\tau_i$, so useful and redundant actions at those steps receive the same learning signal.
To distinguish these actions, we keep this outcome term and add a process advantage that is standardized over all steps in $\mathcal{B}_g$, following StainFlow~\cite{hao2026stainflow}. Let $\mu^{\mathrm{pr}}$ and $\sigma^{\mathrm{pr}}$ be the mean and standard deviation of $\{r_{i,t}^{\mathrm{pr}}\}$ across every step of every trajectory in the sampled group, and let $\alpha>0$ be a coefficient:
\begin{equation}
    \hat{A}_{i,t}^{\mathrm{pr}}
    = \frac{r_{i,t}^{\mathrm{pr}}-\mu^{\mathrm{pr}}}{\sigma^{\mathrm{pr}}+\epsilon},
    \qquad
    A_{i,t}
    = \hat{A}_i^{\mathrm{out}} + \alpha\hat{A}_{i,t}^{\mathrm{pr}}.
\end{equation}
Policy optimization uses $A_{i,t}$ in place of the uniform outcome advantage.
\section{Experiments}
\label{sec:experiments}
\begin{table}[t]
\caption{Process reward evaluation on CUARewardBench. Official baselines use the OpenCUA-Reflector template. P, NPV, R, S, and OA denote precision, negative predictive value, recall, specificity, and overall accuracy. Best results are in bold; second-best are underlined. $\dagger$ indicates our implementations using Qwen3.6-27B.}
\label{tab:cuarewardbench-opencua}
\centering
\small
\begin{tabular}{lccccc}
\toprule
Reward model & P & NPV & R & S & OA \\
\midrule
Qwen2.5-VL-7B & $54.4$ & $49.4$ & $53.8$ & $50.0$ & $52.0$ \\
Qwen2.5-VL-32B & $60.3$ & $64.8$ & $79.8$ & $41.5$ & $61.7$ \\
Qwen2.5-VL-72B & $58.5$ & $64.8$ & $83.0$ & $34.8$ & $60.1$ \\
GLM-4.5V-106B & $64.0$ & $78.5$ & $89.0$ & $44.5$ & $67.9$ \\
Qwen3-VL-32B-T & $68.3$ & $77.3$ & $85.2$ & $56.1$ & $71.4$ \\
Qwen3-VL-235B-T & $66.5$ & $75.0$ & $84.1$ & $53.0$ & $69.4$ \\
GPT-5 & $72.7$ & \underline{$87.8$} & \underline{$92.3$} & \underline{$61.6$} & \underline{$77.7$} \\
GUI-Owl-7B & $64.6$ & $61.8$ & $67.0$ & $59.1$ & $63.3$ \\
GUI-Owl-32B & $61.6$ & $61.5$ & $71.4$ & $50.6$ & $61.6$ \\
Qwen3.5-9B & $59.7$ & $56.7$ & $64.3$ & $51.8$ & $58.4$ \\
Qwen3.6-27B & $62.5$ & $87.0$ & $\mathbf{95.1}$ & $36.6$ & $67.3$ \\
\midrule
GUIDE$^{\dagger}$ & $65.5$ & $\mathbf{89.0}$ & $\mathbf{95.1}$ & $44.5$ & $71.1$ \\
StainFlow$^{\dagger}$ & $63.9$ & $60.1$ & $64.3$ & $59.8$ & $62.1$ \\
\midrule
\multicolumn{6}{l}{\textbf{Ours}} \\
Qwen3.5-9B & \underline{$73.6$} & $65.2$ & $64.3$ & $\mathbf{74.4}$ & $69.1$ \\
Qwen3.6-27B & $\mathbf{78.5}$ & $85.1$ & $88.5$ & \underline{$73.2$} & $\mathbf{81.2}$ \\
\bottomrule
\end{tabular}
\end{table}
\begin{table}[t]
\caption{Process reward evaluation on CUARewardBench, compared with models under the SE-WSM template. Best results are in bold; second-best are underlined.}
\label{tab:cuarewardbench-sewsm}
\centering
\small
\begin{tabular}{lccccc}
\toprule
Reward model & P & NPV & R & S & OA \\
\midrule
Qwen2.5-VL-7B & $56.7$ & $54.2$ & $67.0$ & $43.3$ & $55.8$ \\
Qwen2.5-VL-32B & $65.3$ & $62.4$ & $67.8$ & $59.8$ & $64.0$ \\
Qwen2.5-VL-72B & $58.7$ & $67.1$ & $85.2$ & $33.5$ & $60.7$ \\
GLM-4.5V-106B & $69.5$ & $64.2$ & $66.1$ & $67.7$ & $66.9$ \\
Qwen3-VL-32B-T & $78.2$ & $66.3$ & $63.2$ & \underline{$80.5$} & $71.4$ \\
Qwen3-VL-235B-T & $74.5$ & $63.2$ & $59.3$ & $77.4$ & $67.9$ \\
GPT-5 & $\mathbf{83.7}$ & \underline{$68.8$} & $64.8$ & $\mathbf{86.0}$ & \underline{$74.9$} \\
GUI-Owl-7B & $56.6$ & $64.2$ & $86.8$ & $26.2$ & $58.1$ \\
GUI-Owl-32B & $57.4$ & \underline{$68.8$} & $\mathbf{89.0}$ & $26.8$ & $59.5$ \\
SE-WSM-7B & $58.7$ & $52.0$ & $48.4$ & $62.2$ & $54.9$ \\
Qwen3.5-9B & $60.1$ & $53.7$ & $52.2$ & $61.6$ & $56.6$ \\
Qwen3.6-27B & $73.4$ & $68.2$ & $69.8$ & $72.0$ & $70.8$ \\
\midrule
\multicolumn{6}{l}{\textbf{Ours}} \\
Qwen3.5-9B & $73.6$ & $65.2$ & $64.3$ & $74.4$ & $69.1$ \\
Qwen3.6-27B & \underline{$78.5$} & $\mathbf{85.1}$ & \underline{$88.5$} & $73.2$ & $\mathbf{81.2}$ \\
\bottomrule
\end{tabular}
\end{table}
\begin{table}[t]
\caption{Step-level evaluation on CUAStepBench. Best results are in bold. $\dagger$ indicates results obtained using our implementations.}
\label{tab:cuastepbench-prm}
\centering
\small
\begin{tabular}{llccc}
\toprule
Framework & Model & P & R & F1 \\
\midrule
CUAJudge & GPT-5-mini & $49.7$ & $30.8$ & $38.0$ \\
CUAJudge & Qwen3-VL-8B & $\mathbf{67.7}$ & $12.7$ & $21.3$ \\
OSThemis & Qwen3-VL-8B & $17.0$ & $14.3$ & $15.6$ \\
SeekJudge & Qwen3-VL-8B & $41.4$ & $20.4$ & $27.3$ \\
SeekJudge & SeekJudge-9B & $44.5$ & $33.3$ & $38.1$ \\
SeekJudge & Qwen3.5-9B & $42.5$ & $24.9$ & $31.4$ \\
SeekJudge & Qwen3.6-27B & $44.0$ & $30.8$ & $36.3$ \\
\midrule
GUIDE$^{\dagger}$ & Qwen3.6-27B & $22.9$ & $52.4$ & $31.9$ \\
StainFlow$^{\dagger}$ & Qwen3.6-27B & $11.2$ & $\mathbf{88.7}$ & $20.0$ \\
\midrule
\multirow{2}{*}{\textbf{Ours}} & Qwen3.5-9B & $28.0$ & $63.9$ & $38.9$ \\
 & Qwen3.6-27B & $30.9$ & $57.0$ & $\mathbf{40.1}$ \\
\bottomrule
\end{tabular}
\end{table}
\begin{table*}[t]
\caption{Task success rate (\%) on OSWorld. All methods use UI-TARS-1.5-7B as the base model. Impress and Calc denote LibreOffice Impress and LibreOffice Calc. OA is the overall success rate over the three applications' tasks. Three-run results are reported as mean $\pm$ standard deviation.}
\label{tab:osworld-rl}
\centering
\setlength{\tabcolsep}{3.5pt}
\begin{tabular}{lcccccc@{\hspace{1.2em}}cc}
\toprule
& \multicolumn{2}{c}{Chrome} & \multicolumn{2}{c}{Impress} & \multicolumn{2}{c}{Calc} & \multicolumn{2}{@{\hspace{1.2em}}c}{OA} \\
\cmidrule(lr){2-3} \cmidrule(lr){4-5} \cmidrule(lr){6-7} \cmidrule(l){8-9}
Method & 1 run & 3-run mean & 1 run & 3-run mean & 1 run & 3-run mean & 1 run & 3-run mean \\
\midrule
UI-TARS & $32.6$ & $28.2\pm 4.39$ & $33.8$ & $32.1\pm 2.19$ & $10.6$ & $7.6\pm 2.66$ & $25.6$ & $22.6\pm 2.93$ \\
Outcome GRPO & $43.5$ & $37.0\pm 6.52$ & $44.7$ & $41.6\pm 3.45$ & $19.1$ & $12.8\pm 5.63$ & $35.7$ & $30.4\pm 5.09$ \\
SeekJudge & $45.7$ & $38.3\pm 6.76$ & $46.8$ & $41.1\pm 5.64$ & $19.1$ & $14.9\pm 4.26$ & $37.1$ & $31.4\pm 5.52$ \\
Ours & $\mathbf{49.9}$ & $\mathbf{43.4\pm 6.48}$ & $\mathbf{48.7}$ & $\mathbf{43.3\pm 5.18}$ & $\mathbf{23.4}$ & $\mathbf{17.0\pm 5.63}$ & $\mathbf{40.6}$ & $\mathbf{34.5\pm 5.70}$ \\
\bottomrule
\end{tabular}
\end{table*}
\subsection{Experimental Setup}
\label{sec:exp-setup}
\textbf{Benchmarks.}
We evaluate the proposed process rewards on two process-reward benchmarks and for reinforcement learning on OSWorld applications.
CUARewardBench~\cite{lin2026cuarewardbench} is built on OSWorld trajectories from 10 desktop software categories and 7 agents; it annotates key good and bad actions, and we report process-reward results compared to the benchmark's OpenCUA-Reflector~\cite{wang2025opencuaopenfoundationscomputeruse} and SE-WSM~\cite{sun2025seagent} prompt templates.
CUAStepBench~\cite{wan2026seekjudge} densely annotates every step on executed computer-use trajectories and pairs these labels with human trajectory-level verdicts, covering 177 applications across platforms.
OSWorld~\cite{xie2025osworld} provides open-ended GUI tasks in a real desktop environment, with success judged by execution-based scripts on the final state. We conduct training separately for its Chrome, LibreOffice Impress, and LibreOffice Calc applications. Following the task-selection protocol of DART~\cite{li2025dart}, we select $75$ training tasks in total based on whether UI-TARS-1.5-7B can solve them under Pass@32. Evaluation is conducted separately on all tasks in each application, totaling $140$ tasks across the three applications, including $65$ held-out tasks.
\textbf{Metrics.}
CUARewardBench treats each step judgment as a binary decision against a human label. For our method, we use a threshold of $0.5$ to separate positive and negative predictions. We follow the benchmark protocol: precision (P) and negative predictive value (NPV) measure how reliable a positive or negative verdict is, while recall (R), specificity (S), and overall accuracy (OA) summarize coverage and overall correctness.
On CUAStepBench, every step carries a binary human label, and we score the predicted step judgments by precision (P), recall (R), and F1.
OSWorld’s execution-based evaluators assign each trajectory a score in [0,1]; we report a single-run success rate, as well as the mean and standard deviation across three repeated evaluations of the same checkpoint.
\textbf{Implementation.}
We use Qwen3.6-27B~\cite{qwen2026qwen36} for $F_{\mathrm{part}}$, $F_{\mathrm{ref}}$, $F_{\mathrm{act}}$, $F_{\mathrm{grp}}$, and $F_{\mathrm{rev}}$. We set the aggregation factor to $\epsilon_{\mathrm{g}}=0.1$.
We build on the open-source DART-GUI infrastructure~\cite{li2025dart} for reinforcement learning: OSWorld rollouts are collected independently of training, stored, and consumed by an asynchronous GRPO trainer. Our framework for process-reward generation is integrated as an additional asynchronous stage on completed trajectories, and the trainer uses these process rewards to shape the GRPO advantage of the base model UI-TARS-1.5-7B~\cite{qin2025ui}.
Rollouts use $N=8$ trajectories per instruction, 15 concurrent environments, a maximum of 30 environment steps, and sampling temperature $0.7$; each action is generated with at most 5 images and 35 text turns of history. GRPO training uses learning rate $1\times 10^{-7}$, KL coefficient $0.1$, clip range $[0.1,0.28]$, and process-advantage coefficient $\alpha=0.5$. We train for 30 iterations and evaluate the final checkpoint.
\textbf{Baselines.}
On CUARewardBench, we compare against the official baselines: general vision-language models (Qwen2.5-VL~\cite{bai2025qwen25vl}, Qwen3-VL~\cite{bai2025qwen3vl}, GLM-4.5V~\cite{hong2025glm45v}, and GPT-5~\cite{openai2025gpt5}) and GUI models (GUI-Owl~\cite{ye2025guiowl} and SE-WSM~\cite{sun2025seagent}). We also run the same Qwen3.6-27B and Qwen3.5-9B under the official templates.
On CUAStepBench, we compare against CUAJudge~\cite{xue2026cuajudge}, OSThemis~\cite{li2026osthemis}, and SeekJudge~\cite{wan2026seekjudge} with both its specialist SeekJudge-9B model trained with step-level supervision and the same Qwen3.6-27B backbone.
On both benchmarks, we compare with our implementations of StainFlow~\cite{hao2026stainflow} and GUIDE~\cite{zhai2026guide}, both using Qwen3.6-27B. For GUIDE, we extend its subtask diagnoses to produce binary judgments for each action, enabling comparison on the two process-reward benchmarks. For StainFlow, we use the original prompts released with the paper for entity extraction, observation analysis, key-node verification, and completion assessment.
For reinforcement learning on OSWorld, we compare against the base model, outcome-only GRPO~\cite{shao2024deepseekmath}, and SeekJudge as a process-reward alternative.
\subsection{Process Reward Evaluation}
\label{sec:exp-prm}
Tables~\ref{tab:cuarewardbench-opencua} and~\ref{tab:cuarewardbench-sewsm} compare the proposed framework with official CUARewardBench baselines run under the OpenCUA-Reflector and SE-WSM templates, respectively. Table~\ref{tab:cuastepbench-prm} reports precision, recall, and F1 for identifying error steps on CUAStepBench.
As shown in Tables~\ref{tab:cuarewardbench-opencua} and~\ref{tab:cuarewardbench-sewsm}, our framework achieves the highest overall accuracy ($81.2$), $3.5$ points above GPT-5 with OpenCUA-Reflector ($77.7$) and $6.3$ points above GPT-5 with SE-WSM ($74.9$). Our framework also improves overall accuracy by $13.9$ points over the same Qwen3.6-27B backbone under OpenCUA-Reflector ($67.3$) and by $10.4$ points under SE-WSM ($70.8$), demonstrating that the gain comes from the grouping and two-level scoring rather than from a stronger judge model. The two templates also differ: OpenCUA-Reflector gives higher recall but lower specificity, and SE-WSM gives higher specificity but lower recall (e.g., Qwen3.6-27B at $95.1$/$36.6$ vs.\ $69.8$/$72.0$; GPT-5 at $92.3$/$61.6$ vs.\ $64.8$/$86.0$), while our framework keeps a more even trade-off ($88.5$ recall, $73.2$ specificity). With the same Qwen3.6-27B backbone, our framework exceeds GUIDE ($71.1$) and StainFlow ($62.1$) by $10.1$ and $19.1$ OA points, respectively. The gain over GUIDE supports the effectiveness of our group-based two-level reward assessment over adapting subtask diagnoses to action-level evaluation. With Qwen3.5-9B, our framework achieves $69.1$ OA, exceeding OpenCUA-Reflector and SE-WSM with the same backbone by $10.7$ and $12.5$ points, respectively. It also outperforms the other models of comparable size ($7$--$9$B) under the official templates in both tables.
We then evaluate our framework on CUAStepBench, as shown in Table~\ref{tab:cuastepbench-prm}. With Qwen3.6-27B, our framework achieves the highest F1 of $40.1$, $2.0$ points above SeekJudge-9B ($38.1$) and $3.8$ points above SeekJudge with the same backbone ($36.3$). Compared with SeekJudge using the same backbone, our framework increases recall from $30.8$ to $57.0$ at lower precision ($30.9$ vs.\ $44.0$), thus reducing missed errors that could otherwise receive positive process rewards and reinforce incorrect actions. Our framework also outperforms GUIDE ($31.9$) and StainFlow ($20.0$) by $8.2$ and $20.1$ F1 points, respectively. With Qwen3.5-9B, our framework achieves $38.9$ F1, exceeding SeekJudge with the same backbone ($31.4$) by $7.5$ points. It also exceeds the similarly sized specialist SeekJudge-9B ($38.1$) by $0.8$ F1 points.
\subsection{Reinforcement Learning on OSWorld}
\label{sec:exp-osworld}
As shown in Table~\ref{tab:osworld-rl}, we evaluate RL-trained UI-TARS-1.5-7B on all tasks in OSWorld's Chrome, LibreOffice Impress, and Calc. UI-TARS is the base model without RL; Outcome GRPO is standard GRPO with the outcome reward; SeekJudge and Ours augment GRPO with process rewards from SeekJudge and our framework, respectively. Ours achieves the highest success rate on all three applications and the highest overall success rate (OA). OA exceeds UI-TARS by $15.0$/$11.9$ points and Outcome GRPO by $4.9$/$4.1$ (1-run / 3-run mean). This shows that process-reward advantage shaping improves over outcome-only GRPO. With the same shaping, Ours further outperforms SeekJudge by $3.5$/$3.1$ OA points. Results on the training and held-out subsets and RL ablations are in Appendix~\ref{app:osworld-experiments}.
\subsection{Ablation Study}
\label{sec:exp-ablation}
As shown in Table~\ref{tab:ablation-grouping}, we ablate grouping on CUARewardBench. Ours uses adaptive grouping from the instruction and the agent's responses. Single-group places all actions in one group. Fixed-window denotes the fixed-window step-reward methods described in \secref{sec:intro}, using consecutive windows of 5 screenshots. Grouping w/ screenshots uses screenshots as extra input when partitioning. Ours achieves the highest overall accuracy ($81.2$). Single-group lowers overall accuracy to $75.4$, which shows that splitting the trajectory into subgoal groups is useful because a subgoal is specific enough to tell useful actions from redundant or incorrect ones, rather than judging them against the whole task. Fixed-window lowers overall accuracy further to $73.4$, since a preset local window can miss evidence outside the window or bring in context that does not belong to the executed subgoal. Grouping w/ screenshots also hurts, bringing overall accuracy down to $80.1$, because screenshots can introduce interference, as visually similar screenshots may belong to different subgoals. As shown in Table~\ref{tab:ablation-scoring}, we ablate scoring on CUARewardBench. w/o reference removes $u_k$. w/o refinement removes $F_{\mathrm{rev}}$ and keeps the initial group-level scores. w/o group scoring removes $F_{\mathrm{grp}}$ and $F_{\mathrm{rev}}$ and uses only $r_t^{\mathrm{act}}$. w/o action scoring removes $F_{\mathrm{act}}$ and uses only $\tilde{r}_{k(t)}^{\mathrm{group}}$. Ours achieves the highest overall accuracy, while removing any scoring stage reduces performance, demonstrating the effectiveness of each stage. Additional reward aggregation ablations are in Appendix~\ref{app:aggregation}.
\begin{table}[t]
\caption{Grouping ablation on CUARewardBench. Single-group places all actions in one group. Fixed-window denotes the fixed-window step-reward methods, consecutive windows of 5 screenshots. Grouping w/ screenshots additionally uses screenshots when partitioning.}
\label{tab:ablation-grouping}
\centering
\small
\setlength{\tabcolsep}{3pt}
\begin{tabular}{@{}lccccc@{}}
\toprule
Grouping & P & NPV & R & S & OA \\
\midrule
Single-group & $68.6$ & $\mathbf{96.5}$ & $\mathbf{98.4}$ & $50.0$ & $75.4$ \\
Fixed-window & $67.0$ & $93.9$ & $97.3$ & $47.0$ & $73.4$ \\
Grouping w/ screenshots & $\mathbf{78.7}$ & $81.9$ & $85.2$ & $\mathbf{74.4}$ & $80.1$ \\
Ours & $78.5$ & $85.1$ & $88.5$ & $73.2$ & $\mathbf{81.2}$ \\
\bottomrule
\end{tabular}
\end{table}
\begin{table}[t]
\caption{Scoring ablation on CUARewardBench. w/o reference omits $u_k$. w/o refinement omits $F_{\mathrm{rev}}$. w/o group scoring uses only $r_t^{\mathrm{act}}$. w/o action scoring uses only $\tilde{r}_{k(t)}^{\mathrm{group}}$.}
\label{tab:ablation-scoring}
\centering
\small
\begin{tabular}{lccccc}
\toprule
Scoring & P & NPV & R & S & OA \\
\midrule
w/o reference & $73.0$ & $83.9$ & $89.0$ & $63.4$ & $76.9$ \\
w/o action scoring & $67.7$ & $85.7$ & $92.3$ & $51.2$ & $72.8$ \\
w/o group scoring & $70.2$ & $\mathbf{91.8}$ & $\mathbf{95.6}$ & $54.9$ & $76.3$ \\
w/o refinement & $74.7$ & $86.4$ & $90.7$ & $65.9$ & $78.9$ \\
Ours & $\mathbf{78.5}$ & $85.1$ & $88.5$ & $\mathbf{73.2}$ & $\mathbf{81.2}$ \\
\bottomrule
\end{tabular}
\end{table}
\subsection{User Study on Grouping Quality}
\label{sec:exp-grouping-study}
We assess the quality of $100$ groups produced by our framework on CUARewardBench, sampled across applications. An MLLM-based grouping quality evaluator (\texttt{gpt-6.1-sol}) and two human annotators each rate the same $100$ groups. Both examine the task instruction, the target group's description and action boundaries, its actions and screenshots, and the trajectory context. They rate two dimensions: (1) \emph{Coherence}, assessing whether the actions form a coherent subgoal with natural boundaries; and (2) \emph{Alignment}, assessing whether the group description matches the actions' intended purpose. Each dimension is rated from $1$ (unusable) to $5$ (fully appropriate), with $2$, $3$, and $4$ indicating major issues, a need for revision, and minor issues, respectively. Table~\ref{tab:grouping-user-study} reports these ratings, which support the quality and reliability of the evaluated groupings as contexts for process reward evaluation. A qualitative analysis of the process rewards is given in Appendix~\ref{app:qual}.
\begin{table}[t]
\caption{Grouping quality assessment on 100 groups rated by both human annotators and the MLLM. Human mean averages the two annotators' ratings.}
\label{tab:grouping-user-study}
\centering
\small
\setlength{\tabcolsep}{3pt}
\begin{tabular}{@{}lcccc@{}}
\toprule
\begin{tabular}[c]{@{}l@{}}Dimension\end{tabular} & \begin{tabular}[c]{@{}c@{}}Human\\mean\end{tabular} & \begin{tabular}[c]{@{}c@{}}Human--human\\MAE$\downarrow$\end{tabular} & \begin{tabular}[c]{@{}c@{}}MLLM\end{tabular} & \begin{tabular}[c]{@{}c@{}}Human--MLLM\\MAE$\downarrow$\end{tabular} \\
\midrule
Coherence & $4.94$ & $0.08$ & $4.94$ & $0.06$ \\
Alignment & $4.93$ & $0.10$ & $4.86$ & $0.07$ \\
\bottomrule
\end{tabular}
\end{table}
\subsection{Efficiency Analysis}
\label{sec:efficiency}
As shown in Table~\ref{tab:efficiency-rl}, we compare sampling throughput during OSWorld RL with and without our framework, under synchronous and asynchronous scoring. Under the synchronous setting, the next trajectory starts only after process-reward scoring of the current one has finished. Under the asynchronous setting, the next trajectory starts immediately, and the completed trajectory is scored in the background. Sampling without our framework reaches $92.3$ trajectories per hour under both settings. Directly adding our framework in the synchronous setting drops throughput to $60.3$, whereas our asynchronous setting slightly reduces it to $87.9$. Critic inference costs and performance comparisons at matched inference cost are reported in Appendix~\ref{app:inference-cost}.
\begin{table}[t]
\caption{Sampling throughput (trajectories per hour). Sync starts the next trajectory after scoring finishes; Async starts the next immediately and scores the completed trajectory in the background. $\Delta$ is the change from adding Ours.}
\label{tab:efficiency-rl}
\centering
\small
\begin{tabular}{lcc}
\toprule
Setting & Sync & Async \\
\midrule
w/o Ours & $\mathbf{92.3}$ & $\mathbf{92.3}$ \\
Ours & $60.3$ & $87.9$ \\
\midrule
$\Delta$ & $-32.0$ & $-4.4$ \\
\bottomrule
\end{tabular}
\end{table}
\section{Conclusion}
\label{sec:conclusion}
In this paper, we present an adaptive-group process reward framework that dynamically partitions a GUI trajectory into semantically coherent subgoal groups, enabling process reward estimation at both the action and group levels. The process reward of each action is obtained by combining its action-level evaluation with the refined group-level evaluation. These process rewards are used to shape GRPO's step-level advantage during policy optimization. Experiments on process reward benchmarks and downstream RL training demonstrate the effectiveness of the proposed framework.

\section*{LLM Usage Statement}
We used OpenAI (ChatGPT / GPT) and Google (Gemini). Major parts of the manuscript were given to these providers for reviewer-style feedback. For writing and editing, related work and citations, the design, execution, and analysis of experiments, figures and diagrams, and the wording of research ideas, hypotheses, claims, and conclusions, the tools only edited or refined author-produced content. For code and implementation, they generated material that the authors reviewed, edited, and approved. For data, including annotation, they generated the ratings in the user study, which the authors reviewed and approved. The authors checked all affected material and are responsible for the final paper.
{
    \small
    \bibliographystyle{ieeenat_fullname}
    \bibliography{references}
}
\clearpage
\maketitlesupplementary

\appendix
\section{Reward Aggregation Ablation}
\label{app:aggregation}
We compare different values of $\epsilon_{\mathrm{g}}$ in process reward aggregation on CUARewardBench with a fixed decision threshold of $\delta=0.5$. Setting $\epsilon_{\mathrm{g}}=0$ directly multiplies the action-level and refined group-level scores. We also evaluate a weighted sum, $r_t^{\mathrm{pr}}=\lambda r_t^{\mathrm{act}}+(1-\lambda)\tilde{r}_{k(t)}^{\mathrm{group}}$, with $\lambda=0.5$ and $\delta=0.8$, selected to maximize OA in the threshold sweep.. Results are reported in Table~\ref{tab:aggregation-ablation}. The weighted sum achieves similar accuracy to our formulation. Both combine the same action-level and group-level assessments, suggesting that the two aggregation rules can yield similar classification performance after adjusting for their different score scales.
\begin{table}[t]
\caption{Reward aggregation ablation on CUARewardBench.}
\label{tab:aggregation-ablation}
\centering
\small
\setlength{\tabcolsep}{3pt}
\begin{tabular}{@{}l|cc|ccccc@{}}
\toprule
Aggregation & $\epsilon_{\mathrm{g}}$ & $\delta$ & P & NPV & R & S & OA \\
\midrule
\multirow{5}{*}{Product} & $0$ & $0.5$ & $78.4$ & $84.5$ & $87.9$ & $\mathbf{73.2}$ & $80.9$ \\
 & $0.1$ & $0.5$ & $\mathbf{78.5}$ & $85.1$ & $88.5$ & $\mathbf{73.2}$ & $\mathbf{81.2}$ \\
 & $0.2$ & $0.5$ & $76.3$ & $86.3$ & $90.1$ & $68.9$ & $80.1$ \\
 & $0.3$ & $0.5$ & $76.3$ & $86.3$ & $90.1$ & $68.9$ & $80.1$ \\
 & $0.4$ & $0.5$ & $74.9$ & $\mathbf{87.8}$ & $\mathbf{91.8}$ & $65.9$ & $79.5$ \\
\midrule
Weighted sum & --- & $0.8$ & $78.4$ & $84.5$ & $87.9$ & $\mathbf{73.2}$ & $80.9$ \\
\bottomrule
\end{tabular}
\end{table}
\section{Inference Cost Analysis}
\label{app:inference-cost}
To assess the inference cost of generating process rewards, we report total token usage, token usage per input trajectory, and throughput. CUARewardBench measures reward accuracy on annotated key steps; for a fair cost comparison, we apply every method to all action steps in each trajectory. All compared methods use Qwen3.6-27B on CUARewardBench and Qwen3.5-9B on CUAStepBench.
As shown in Table~\ref{tab:inference-cost-cuarewardbench}, OpenCUA uses $2.71\times$ as many tokens as our framework, while our framework achieves $1.46\times$ its throughput. Compared with SE-WSM, our framework uses $2.66\times$ as many tokens and achieves $0.28\times$ its throughput. On CUAStepBench (Table~\ref{tab:inference-cost-cuastepbench}), SeekJudge uses $1.65\times$ as many tokens as our framework, while our framework achieves $1.70\times$ its throughput.
\begin{table*}[t]
\caption{Critic inference cost on CUARewardBench. OpenCUA denotes OpenCUA-Reflector.}
\label{tab:inference-cost-cuarewardbench}
\centering
\small
\setlength{\tabcolsep}{6pt}
\begin{tabular}{@{}lrrr@{}}
\toprule
Method & Total tokens (M) & Tokens / trajectory (K) & Throughput (trajectories/h) \\
\midrule
SE-WSM & 7.18 & 26.41 & 243.88 \\
OpenCUA & 51.77 & 190.33 & 46.49 \\
\midrule
Ours & 19.09 & 70.19 & 67.89 \\
\bottomrule
\end{tabular}
\end{table*}
\begin{table*}[t]
\caption{Critic inference cost on CUAStepBench.}
\label{tab:inference-cost-cuastepbench}
\centering
\small
\setlength{\tabcolsep}{8pt}
\begin{tabular}{@{}lrrr@{}}
\toprule
Method & Total tokens (M) & Tokens / trajectory (K) & Throughput (trajectories/h) \\
\midrule
SeekJudge & 56.88 & 204.59 & 51.77 \\
\midrule
Ours & 34.53 & 124.19 & 88.17 \\
\bottomrule
\end{tabular}
\end{table*}
We further compare our framework with SE-WSM at comparable inference cost on CUARewardBench. To increase the inference cost of SE-WSM toward that of our framework, we perform 3 independent scoring passes using its original template and combine the resulting action-level judgments by majority vote. Table~\ref{tab:sewsm-cost-comparison} reports the token usage and performance of each setting. Majority voting improves SE-WSM's OA from $70.8$ to $73.4$, while our framework achieves higher scores on all reported metrics and exceeds its OA by $7.8$ percentage points ($81.2$). These results suggest that the gains of our framework are not solely attributable to increased inference cost.
\begin{table*}[t]
\caption{Performance comparison with SE-WSM at comparable inference cost on CUARewardBench. All methods use Qwen3.6-27B.}
\label{tab:sewsm-cost-comparison}
\centering
\small
\setlength{\tabcolsep}{4pt}
\begin{tabular}{@{}lcccccc@{}}
\toprule
Method & Tokens / trajectory (K) & P & NPV & R & S & OA \\
\midrule
SE-WSM & 26.41 & $73.4$ & $68.2$ & $69.8$ & $72.0$ & $70.8$ \\
SE-WSM + voting & 77.59 & $74.7$ & $72.0$ & $74.7$ & $72.0$ & $73.4$ \\
\midrule
Ours & 70.19 & $78.5$ & $85.1$ & $88.5$ & $73.2$ & $81.2$ \\
\bottomrule
\end{tabular}
\end{table*}
\section{Additional Experiments on OSWorld}
\label{app:osworld-experiments}
\subsection{Training and Held-out Evaluation}
\label{app:osworld-splits}
Training is conducted separately for Chrome, LibreOffice Impress, and LibreOffice Calc. Evaluation covers all tasks in each application. Following the task-selection protocol of DART~\cite{li2025dart}, the $75$ training tasks are selected based on whether UI-TARS-1.5-7B can solve them under Pass@32. Held-out tasks are those excluded from training. Table~\ref{tab:osworld-split-counts} lists the task counts.
\begin{table}[t]
\caption{OSWorld task counts for training and held-out evaluation.}
\label{tab:osworld-split-counts}
\centering
\small
\begin{tabular}{lrrr}
\toprule
Application & Training & Held-out & All \\
\midrule
Chrome & $31$ & $15$ & $46$ \\
Impress & $31$ & $16$ & $47$ \\
Calc & $13$ & $34$ & $47$ \\
\midrule
Total & $75$ & $65$ & $140$ \\
\bottomrule
\end{tabular}
\end{table}
Table~\ref{tab:osworld-split-results} reports results on the training and held-out subsets. Our framework achieves the highest OA on both subsets in the single-run results and three-run means, suggesting that its improvements extend to held-out tasks.
\begin{table*}[t]
\caption{Task success rate (\%) on OSWorld training and held-out tasks. OA is the overall success rate over the three applications' tasks. Three-run results are means.}
\label{tab:osworld-split-results}
\centering
\small
\setlength{\tabcolsep}{3pt}
\begin{tabular}{@{}l*{8}{c}@{}}
\toprule
& \multicolumn{2}{c}{Chrome} & \multicolumn{2}{c}{Impress} & \multicolumn{2}{c}{Calc} & \multicolumn{2}{c}{OA} \\
\cmidrule(lr){2-3}\cmidrule(lr){4-5}\cmidrule(lr){6-7}\cmidrule(lr){8-9}
Method & Train & Held-out & Train & Held-out & Train & Held-out & Train & Held-out \\
\midrule
\multicolumn{9}{l}{\textbf{Single-run results}} \\
UI-TARS & $48.4$ & $0.0$ & $51.3$ & $0.0$ & $38.5$ & $0.0$ & $47.9$ & $0.0$ \\
Outcome GRPO & $54.8$ & $20.0$ & $61.3$ & $12.5$ & $61.5$ & $2.9$ & $58.7$ & $9.2$ \\
SeekJudge & $58.1$ & $20.0$ & $61.3$ & $18.8$ & $61.5$ & $2.9$ & $60.0$ & $10.8$ \\
Ours & $61.2$ & $26.7$ & $64.2$ & $18.8$ & $69.2$ & $5.9$ & $63.8$ & $13.8$ \\
\midrule
\multicolumn{9}{l}{\textbf{Three-run means}} \\
UI-TARS & $41.8$ & $0.0$ & $48.6$ & $0.0$ & $27.3$ & $0.0$ & $42.1$ & $0.0$ \\
Outcome GRPO & $48.4$ & $13.3$ & $59.8$ & $6.3$ & $38.5$ & $2.9$ & $51.4$ & $6.2$ \\
SeekJudge & $50.3$ & $13.3$ & $59.0$ & $6.3$ & $46.2$ & $2.9$ & $53.2$ & $6.2$ \\
Ours & $54.7$ & $20.0$ & $62.3$ & $6.3$ & $46.2$ & $5.9$ & $56.4$ & $9.2$ \\
\bottomrule
\end{tabular}
\end{table*}
\subsection{Ablation Study}
\label{app:osworld-ablation}
We further ablate the grouping strategy for downstream RL by evaluating the Single-group and Fixed-window variants from Table~\ref{tab:ablation-grouping} on OSWorld's LibreOffice Impress tasks. Table~\ref{tab:osworld-grouping-ablation} shows that adaptive grouping achieves the highest single-run and three-run mean performance. These results further support the effectiveness of adaptive grouping in downstream RL and suggest that more accurate process rewards can provide more reliable learning signals for RL training.
\begin{table}[t]
\caption{Grouping ablation for RL on OSWorld's LibreOffice Impress tasks. Results are reported as task success rates (\%). Three-run results are reported as mean $\pm$ standard deviation.}
\label{tab:osworld-grouping-ablation}
\centering
\small
\setlength{\tabcolsep}{6pt}
\begin{tabular}{@{}lcc@{}}
\toprule
Grouping & 1 run & 3-run mean \\
\midrule
Single-group & $44.5$ & $41.6\pm 3.24$ \\
Fixed-window & $40.2$ & $38.1\pm 2.13$ \\
\midrule
Ours & $48.7$ & $43.3\pm 5.18$ \\
\bottomrule
\end{tabular}
\end{table}
\section{Qualitative Analysis}
\label{app:qual}
\subsection{Error Analysis}
\label{app:error}
We analyze $65$ critic error steps on CUARewardBench, which fall into four categories (Table~\ref{tab:error-causes}).
Insufficient application knowledge ($41.5\%$) is the largest share: the critic does not know the correct menu, setting, syntax, or operator, so its reference treats a wrong path as the intended procedure (e.g., Password Manager instead of cookie settings, Insert-Comment instead of View-Notes, or a Calc sheet reference written with ``!'' rather than ``.'').
Visual or interaction misreads ($23.1\%$) are screenshot-level mistakes: a later date read as earlier, a filter chip taken as on, a slider taken as max, or a click that did not register.
Missed instruction constraints ($18.5\%$) are cases where $g$ already specifies the requirement---working directory, numeric target, time range, ``on sale''---but the critic scores the step without enforcing it.
Overestimated group contribution ($16.9\%$): the group-level score still treats a subgoal as helping the task when it does not (e.g., a group that resizes one layer when the task is to resize the whole image).
\begin{table}[h]
\caption{Underlying causes of critic--human disagreements on CUARewardBench key steps.}
\label{tab:error-causes}
\centering
\small
\begin{tabular}{lcc}
\toprule
Cause & Count & Share \\
\midrule
Insufficient application knowledge & $27$ & $41.5\%$ \\
Visual or interaction misread & $15$ & $23.1\%$ \\
Missed instruction constraint & $12$ & $18.5\%$ \\
Overestimated group contribution & $11$ & $16.9\%$ \\
\bottomrule
\end{tabular}
\end{table}
\subsection{Case Studies}
\label{app:cases}
\begin{figure*}[t]
\centering
\includegraphics[width=\textwidth]{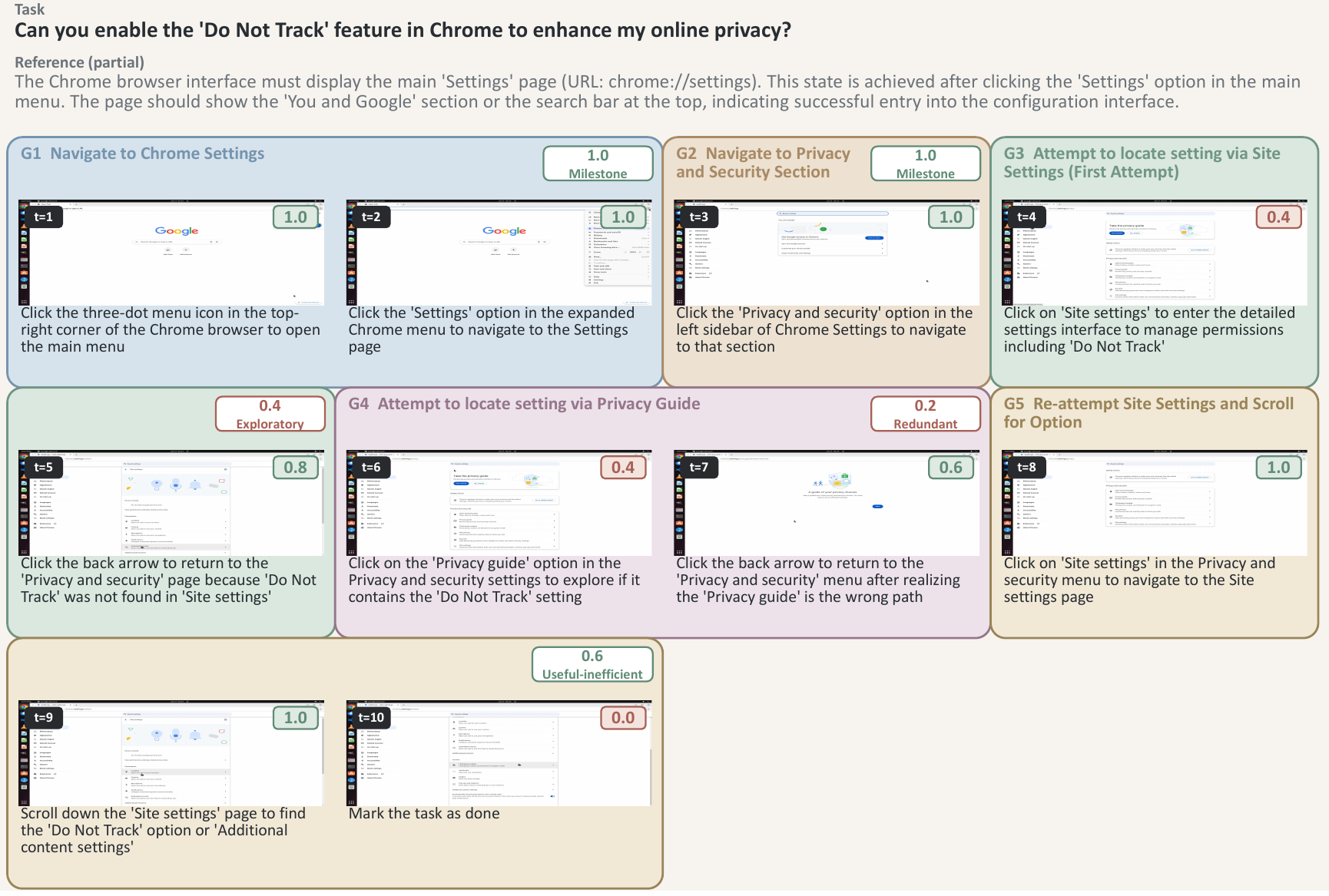}
\caption{A Chrome trajectory for enabling Do Not Track.}
\label{fig:qual-chrome}
\end{figure*}

\begin{figure*}[t]
\centering
\includegraphics[width=\textwidth]{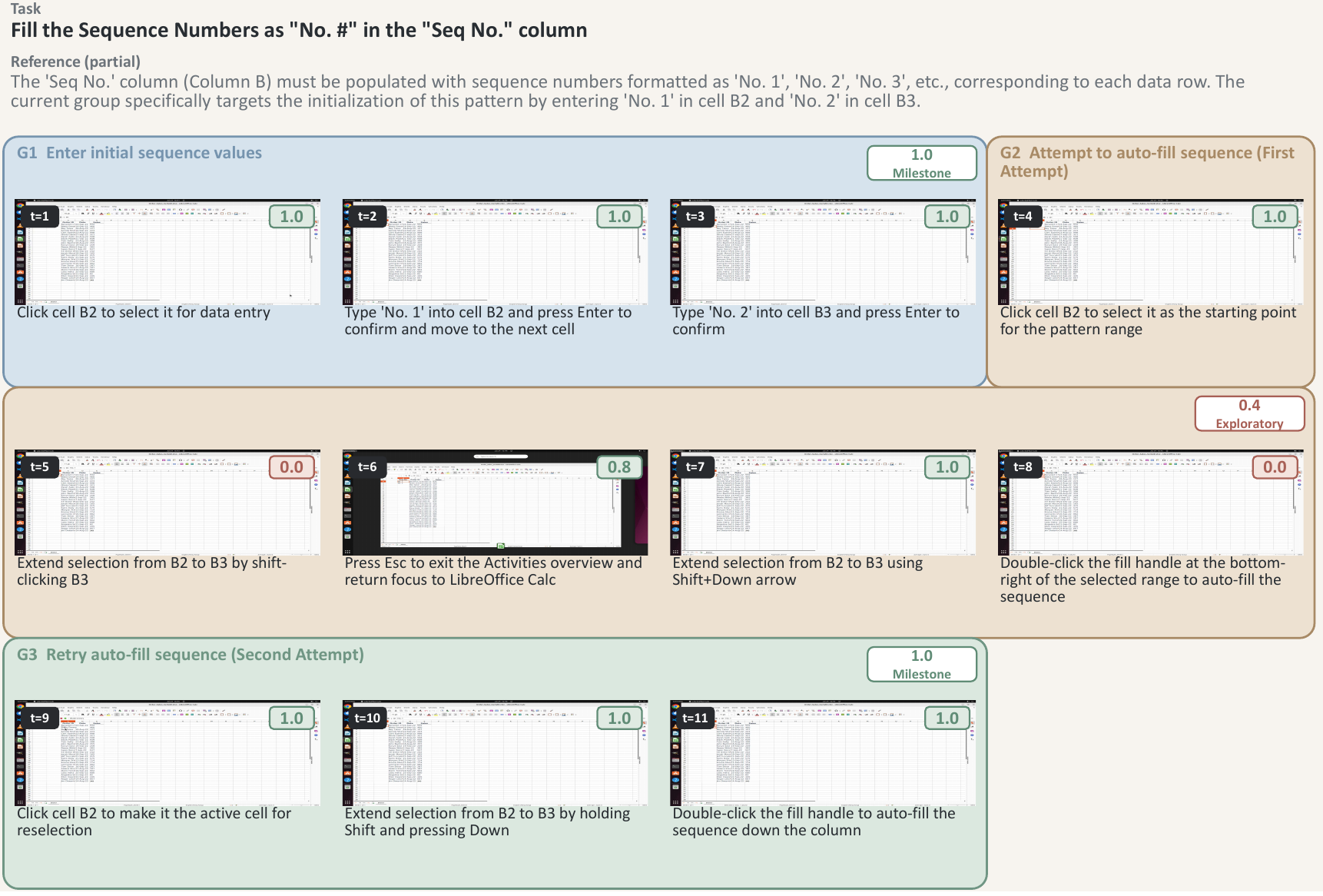}
\caption{A Calc trajectory that fills the Seq No.\ column with ``No.\ \#''.}
\label{fig:qual-calc}
\end{figure*}

\begin{figure*}[t]
\centering
\includegraphics[width=\textwidth]{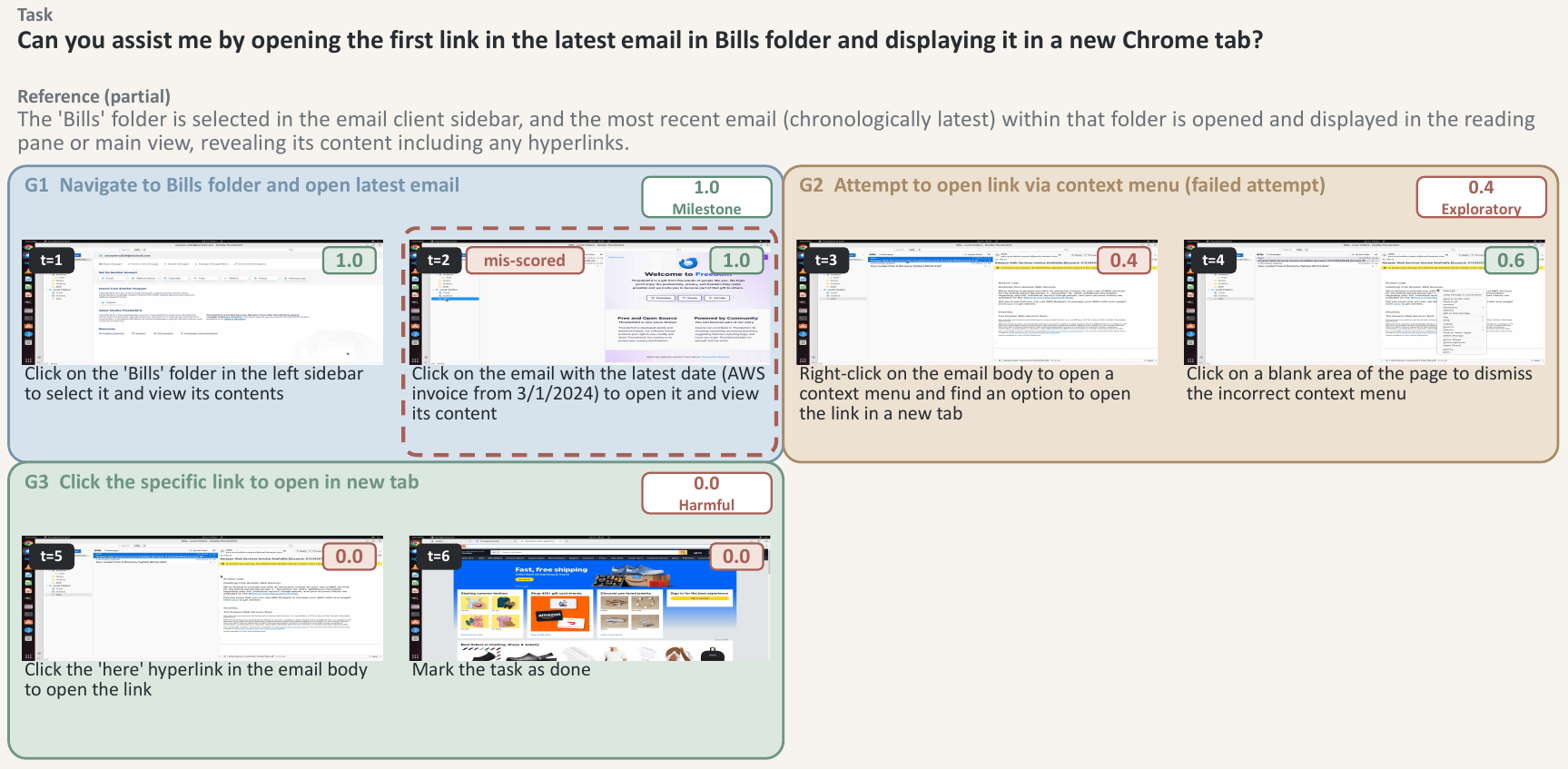}
\caption{A failure-case trajectory that opens the first link in the latest Bills email in a new Chrome tab. The dashed outline marks the incorrectly scored step.}
\label{fig:qual-fail}
\end{figure*}

Figures~\ref{fig:qual-chrome}, \ref{fig:qual-calc}, and \ref{fig:qual-fail} visualize the adaptive-group process rewards on three CUARewardBench trajectories.
Each cell shows the screenshot, the action, and the action-level score.
A colored box groups consecutive steps of the same subgoal and shows the refined group-level score and contribution.
Figure~\ref{fig:qual-chrome} never turns on Do Not Track, but Groups~1--2 (``Navigate to Chrome Settings'' and ``Navigate to Privacy and Security Section'') still score opening Settings and Privacy and security as useful, instead of giving every step a zero because the trajectory fails.
In Group~3, clicking ``Site settings'' is exploratory and the back arrow is Useful-efficient (it undoes the wrong page); Group~4 scores ``Privacy guide'' the same way, then lowers that group to redundant because Group~3 has already shown the setting is not there.
Group~5 scores going back to ``Site settings'' as useful, but the premature \texttt{DONE} as $0$.
Figure~\ref{fig:qual-calc} fills the column only after Group~2 (``Attempt to auto-fill sequence (First Attempt)'') fails.
If only the final outcome were used, the misclick that opens Activities would get the same success credit as the later fill.
Grouping puts typing ``No.\ 1'' and ``No.\ 2'' in Group~1 (``Enter initial sequence values''), the failed fill in Group~2, and the retry in Group~3 (``Retry auto-fill sequence (Second Attempt)''), so the successful retry does not change the scores of the failed attempt.
In Group~2, the click that opens Activities and the missed fill handle are harmful, Esc is Useful-efficient (it closes Activities and returns to Calc), and the group stays exploratory because the column is still empty; Group~3 then scores the same fill as useful after ``No.\ \#'' appears.
Figure~\ref{fig:qual-fail} shows a failure case: the process reward is wrong at the dashed step.
The instruction asks for the latest email in the Bills folder.
The list shows an AWS invoice dated $3/1/2024$ above an X receipt dated $12/1/2024$; $12/1/2024$ is later, so the second mail should be opened.
The critic compares the two dates incorrectly, treats $3/1/2024$ as newer, and scores clicking the AWS invoice as $1.0$.
This is a visual-judgment error limited by the capability of the model used as the critic: it is not reliable enough at reading a small UI detail such as which of two dates is later.
\section{Additional Implementation Details}
\label{app:impl}
We list the score mapping $M$ and the critic prompts for $F_{\mathrm{part}}$, $F_{\mathrm{ref}}$, $F_{\mathrm{act}}$, $F_{\mathrm{grp}}$, and $F_{\mathrm{rev}}$.
\subsection{Score mapping}
\label{app:score-map}
Table~\ref{tab:score-map} maps the two labels to a scalar in $\{0,0.2,0.4,0.6,0.8,1.0\}$.
The same table is used at both levels: $e_t$/$e_k$ as execution or subgoal status, and $p_t$/$p_k$ as subgoal or task contribution.
\begin{table}[h]
\caption{Lookup table $M$ mapping the two labels to a scalar. The same table is used at both the action level and the group level.}
\label{tab:score-map}
\centering
\small
\begin{tabular}{lcccc}
\toprule
Contribution & Completed & Partial & Failed & Uncertain \\
\midrule
Milestone & $1.0$ & $0.4$ & $0.0$ & $0.4$ \\
Useful-efficient & $0.8$ & $0.4$ & $0.0$ & $0.4$ \\
Useful-inefficient & $0.6$ & $0.4$ & $0.0$ & $0.4$ \\
Exploratory & $0.4$ & $0.4$ & $0.0$ & $0.2$ \\
Redundant & $0.2$ & $0.2$ & $0.0$ & $0.2$ \\
Harmful & $0.0$ & $0.0$ & $0.0$ & $0.0$ \\
\bottomrule
\end{tabular}
\end{table}
\subsection{Grouping critic prompt}
\label{app:prompt-part}
$F_{\mathrm{part}}$ reads the instruction $g$ and the responses $\{c_t\}$ and returns contiguous groups $G_k=(n_k,\mathcal{I}_k)$.
\promptlisting{Grouping critic prompt}{prompts/part.txt}
\subsection{Reference critic prompt}
\label{app:prompt-ref}
$F_{\mathrm{ref}}$ builds the instruction-conditioned reference $u_k=(u_k^{\mathrm{tgt}},u_k^{\mathrm{cst}},u_k^{\mathrm{prc}})$ for group $G_k$.
\promptlisting{Reference critic prompt}{prompts/ref.txt}
\subsection{Action-level scoring critic prompt}
\label{app:prompt-act}
$F_{\mathrm{act}}$ scores all target actions of a sliding window inside $G_k$ in one call.
Screenshots are attached only to the window being scored; other actions in the group are given as text.
\promptlisting{Action-level scoring critic prompt}{prompts/act.txt}
\subsection{Group-level scoring critic prompt}
\label{app:prompt-grp}
$F_{\mathrm{grp}}$ scores $G_k$ from the action-level judgments $J_k$, the reference $u_k$, and preceding summaries $H_{<k}$.
\promptlisting{Group-level scoring critic prompt}{prompts/grp.txt}
\subsection{Refinement critic prompt}
\label{app:prompt-rev}
$F_{\mathrm{rev}}$ revisits all group-level labels after every group has been scored, using the full sequence of group conclusions.
\promptlisting{Refinement critic prompt}{prompts/rev.txt}

\end{document}